\documentclass[journal]{IEEEtran}
\usepackage{soul}
\usepackage{color}
\usepackage{amsmath}
\usepackage{physics}
\usepackage{graphicx}
\usepackage{subcaption}
\usepackage{multirow}
\usepackage{caption}
\usepackage[numbers,sort&compress]{natbib}

\begin{document}
%\title{Physics-Informed Neuro-Evolution (PINE): A Survey and Prospects}
\title{Evolutionary Optimization of Physics-Informed Neural Networks: Evo-PINN Frontiers and Opportunities}

%\author{Authors Anonymous}

\DeclareRobustCommand*{\IEEEauthorrefmark}[1]{%
  \raisebox{0pt}[0pt][0pt]{\textsuperscript{\footnotesize #1}}%
}

\author{\IEEEauthorblockN{Jian Cheng Wong\IEEEauthorrefmark{1}, Abhishek Gupta\IEEEauthorrefmark{2}, Chin Chun Ooi\IEEEauthorrefmark{1}, Pao-Hsiung Chiu\IEEEauthorrefmark{1}, Jiao Liu\IEEEauthorrefmark{3}, and Yew-Soon Ong\IEEEauthorrefmark{1,3}}\\
\IEEEauthorblockA{\IEEEauthorrefmark{1}Agency for Science, Technology and Research (A*STAR), 138634, Singapore}\\
\IEEEauthorblockA{\IEEEauthorrefmark{2}Indian Institute of Technology (IIT) Goa, Ponda, 403401, India}\\
\IEEEauthorblockA{\IEEEauthorrefmark{3}Nanyang Technological University, 639798, Singapore}\\
\thanks{Corresponding author: Jian Cheng Wong (e-mail: wongj@a-star.edu.sg).}}

\maketitle

\begin{abstract}
Deep learning models trained on finite data lack a complete understanding of the physical world. On the other hand, physics-informed neural networks (PINNs) are infused with such knowledge through the incorporation of mathematically expressible laws of nature into their training loss function. By complying with physical laws, PINNs provide advantages over purely data-driven models in limited-data regimes and present as a promising route towards Physical AI. This feature has propelled them to the forefront of scientific machine learning, a domain characterized by scarce and costly data. However, the vision of accurate physics-informed learning comes with significant challenges. This work examines PINNs in terms of model optimization and generalization, shedding light on the need for new algorithmic advances to overcome issues pertaining to the training speed, precision, and generalizability of today's PINN models. Of particular interest are gradient-free evolutionary algorithms (EAs) for optimizing the uniquely complex loss landscapes arising in PINN training. Methods synergizing gradient descent and EAs for discovering bespoke neural architectures and balancing multiple terms in physics-informed learning objectives are positioned as important avenues for future research. Another exciting track is to cast EAs as a meta-learner of generalizable PINN models. To substantiate these proposed avenues, we further highlight results from recent literature to showcase the early success of such approaches in addressing the aforementioned challenges in PINN optimization and generalization.
\end{abstract}

\begin{IEEEkeywords}
Physics-informed neural networks, scientific machine learning, evolutionary optimization, neuroevolution, generalizability.
\end{IEEEkeywords}

% \fi
\IEEEpeerreviewmaketitle

\section{Introduction}

\IEEEPARstart{O}{ver} the past decade, advances in deep learning have garnered increasing attention for their potential to address pressing challenges in science and engineering. The emerging topic of artificial intelligence (AI) for science promises to transform technological frontiers~\cite{carleo2019machine}. For example, learned models have shown the capacity to accelerate the design and discovery of novel materials, photonic devices, proteins, drugs, and renewable energy technologies such as wind turbines. However, unlike conventional applications of AI in vision, speech, or natural language processing where there is a deluge of available data, applications in scientific domains are often characterized by extreme data scarcity. This has raised concerns regarding the veracity of such models when trained only on small data. The cost of acquiring quality data gives rise to a practical bottleneck, leading to naive statistical models which sometimes generate physically inconsistent outputs. These shortcomings become even more consequential in the context of Agentic AI systems~\cite{wong2024llm2fea, ghafarollahi2025automating, lu2024ai}, which are designed to autonomously drive scientific discovery and decision-making. In such systems, a lack of embedded physical understanding by models can lead to flawed reasoning or unsafe actions when interfacing with the real world~\cite{wei2025evolvableconditionaldiffusion, xulooks}.

\begin{figure*}[htbp]
\centering
  \begin{subfigure}[b]{0.325\textwidth}
    \includegraphics[width=\textwidth]{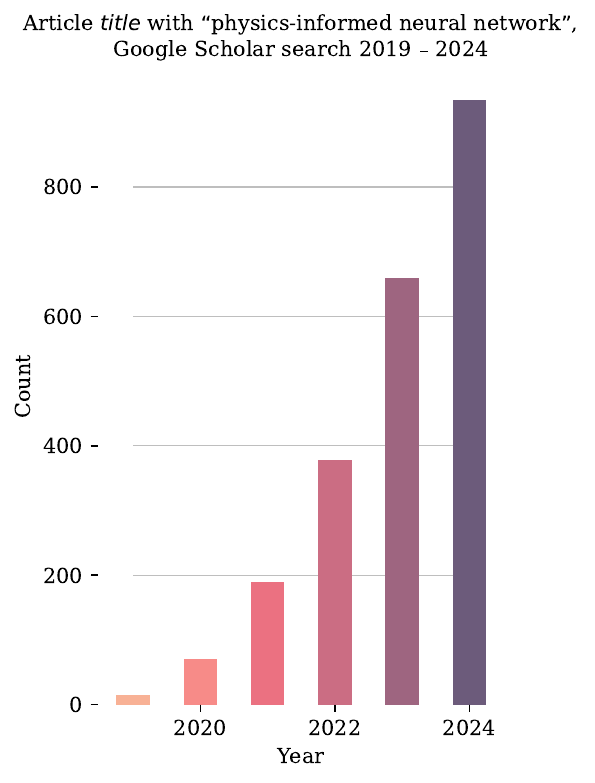}
  \end{subfigure}
  \hfill
  \begin{subfigure}[b]{0.4\textwidth}
    \includegraphics[width=\textwidth]{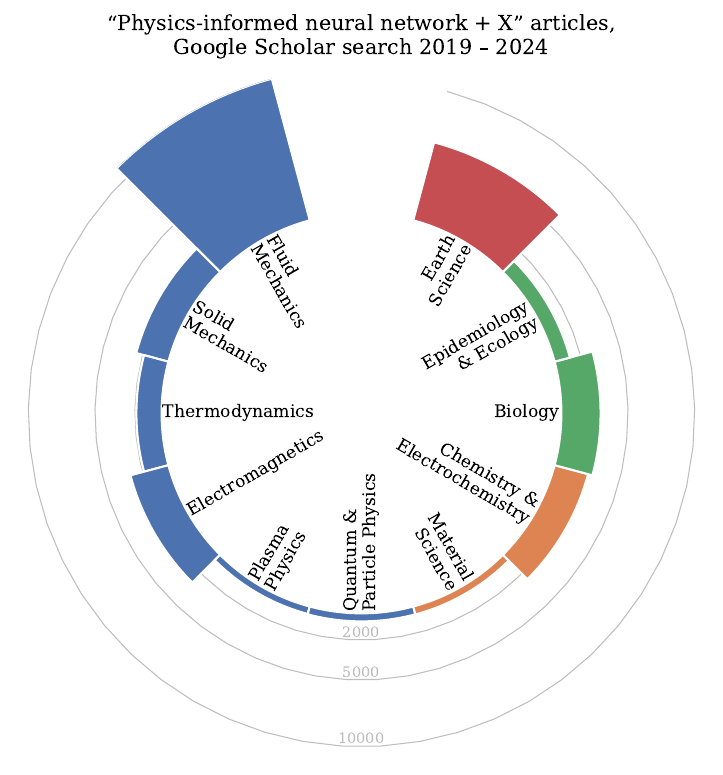}
  \end{subfigure}
  \begin{subfigure}[b]{0.26\textwidth}
    \includegraphics[width=\textwidth]{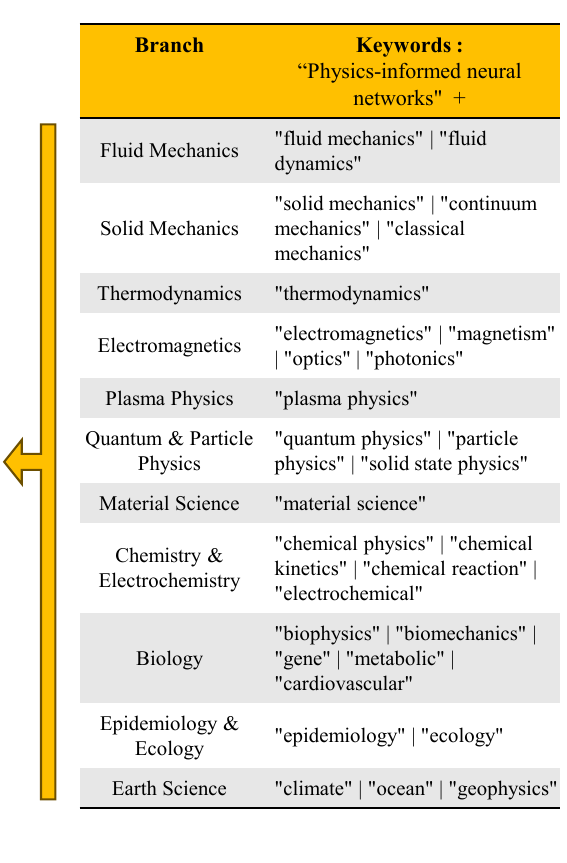}
  \end{subfigure}
  \caption{Google Scholar search of the literature shows the popularity of PINN over the years between 2019 and 2024, and PINN studies in diverse scientific areas. \emph{Data collected in November 2024.}}
\label{fig:pinn-popularity}
\end{figure*}

\begin{table*}[htbp]
\centering
\caption{Potential applications of PINNs across various scientific domains}
\begin{tabular}{l|l}
\hline
Branch & Applications \\
\hline
Fluid Mechanics & Turbomachinery design~\cite{wang2024novel}, airfoil design~\cite{zhang2022aerodynamic} \\
Solid Mechanics & Topology optimization~\cite{maze2023diffusion, huang2024topology}, vehicle crashworthiness design~\cite{sun2014crashing}, additive manufacturing~\cite{xiong2019data, doh2022bayesian}, digital twin~\cite{lai2021designing, he2024digital} \\
{Chemistry \& Electrochemistry} & Battery management systems~\cite{wang2024physics}  \\
{Quantum \& Particle Physics} & Diffuse optical tomography~\cite{zou2021machine}, nuclear fusion physics analysis~\cite{garbet2010gyrokinetic, Garbet_2024} \\
Material Science & Composite-tool systems during manufacture~\cite{niaki2021physics} \\
\hline
\end{tabular}
\label{tab:pinn-applications}
\end{table*}

The shortcomings of standard data-driven models have seeded the field of \emph{scientific machine learning} where minimizing deviations from the known laws of nature serves to boost model precision and data efficiency~\cite{roscher2020explainable,gunning2019darpa}. Associated methods offer unique pathways to \textit{rationalizable} AI, defined in~\cite{ong2019air}, that is cognizant of fundamental physics principles. A myriad of approaches to instil physics consistency into learned models can be crafted, with the most suitable methods being naturally dependent on the task at hand and the type of knowledge available~\cite{karpatne2017theory, von2021informed, willard2022integrating, karniadakis2021physics}. Some instantiations include the design of model architectures that preserve invariances or symmetries representative of the underlying system~\cite{cohen2016group}, data augmentation with theory-based synthetic data (e.g., first principles numerical simulations)~\cite{barbastathis2019use}, or introduction of learning biases into the training process~\cite{battaglia2018relational}.

This effort focuses on the growing body of literature on \textit{physics-informed neural networks} (PINNs), popularized in recent years by the work of Raissi \textit{et al.}~\cite{raissi2019physics}. The essential idea of a PINN---\textit{where mathematically-defined learning biases are directly incorporated into a neural network's loss function}---is simultaneously decades old, with its earlier work in the 1990s~\cite{lee1990neural,lagaris1998artificial}, and also a breath of fresh air for AI. This physics-informed loss is amenable to various forms of scientific knowledge and theories, including conservation laws and/or governing equations in the form of ordinary and partial differential equations (ODEs and PDEs), yet retains immense flexibility in the choice of network architectures (e.g., multilayer perceptron, convolutional or graph neural networks) and learning algorithms (e.g., gradient-based or gradient-free). This makes it possible for scientific knowledge uncovered over centuries, from the micro-scale evolution of quantum-mechanical systems to the macro-scale behavior of solids, fluids, and propagation of electromagnetic fields, to be incorporated into the learned models of today, thereby compensating for a lack of labeled data. % by augmenting deep learning with physics principles.  

%\subsection{The Scientific Influence of PINNs}

The potential of such physics-informed models has spawned significant interest across various scientific domains. Trends from a Google Scholar search of the  literature in the last six years (2019-2024) are depicted in Fig.~\ref{fig:pinn-popularity}. The figure reveals rapid growth in the popularity of PINNs in different branches of science, with many demonstrations being developed using open software libraries such as JAX~\cite{jax2018github}, NVIDIA PhysicsNeMo ~\cite{modulus2023}, SciANN~\cite{haghighat2021sciann}, and DeepXDE~\cite{lu2021deepxde}.

%\subsection{Physics-Informed NeuroEvolution (PINE)}

%\begin{figure*}[htbp]
%\centering
%\includegraphics[width=1.0\linewidth]{Figure/fig-pinn-popularity.pdf}
%\caption{Google Scholar survey of the literature shows (a) the popularity of PINN over the years between 2017 and 2022, and (b) PINN studies in diverse scientific areas.}
%\label{fig:pinn-popularity}
%\end{figure*}

Despite the broad applicability of PINNs (ref. to Table~\ref{tab:pinn-applications} for some potential applications), the switch from standard data-driven loss functions to physics-informed learning objectives has brought unforeseen difficulties in optimizing the latter's uniquely complex landscape~\cite{krishnapriyan2021characterizing, fuks2020limitations, ji2021stiff}. These training difficulties undermine the theoretical advantages of PINNs, and are currently a barrier to their adoption in place of established numerical methods in science and engineering. Prior surveys~\cite{karniadakis2021physics, cuomo2022scientific, cai2021physics, banerjee2024physics, hao2023pinnacle, toscano2025pinns} review PINN methods, applications, limitations, trends, and prospects from interdisciplinary viewpoints, but a common point of emphasis is that efficient optimization of PINNs remains an open problem~\cite{hao2023pinnacle}. %benchmark data, 

Separately, despite the physics-informed loss, the physical validity of a PINN is limited to scenarios encountered during training, even after a complicated training process. Most PINNs are designed to model a specific PDE instance, and model training can be highly sensitive to choice of architecture and optimization hyperparameters. Consequently, reusing identical hyperparameter configurations to train new PINN models within the same PDE family often fails to produce satisfactory results. They do not inherently generalize learned physics to new scenarios beyond their training regime---e.g., changing PDE parameters, initial conditions or boundary conditions---thereby continuing to suffer from generalizability issues similar to standard data-driven learning algorithms. %on similar scenarios 

%This paper shall therefore examine PINNs for the first time through the lens of their optimization and generalization challenges. The goal of fulfilling the vision of scientific machine learning unveils a distinctive potential to adapt computational intelligence techniques developed over the years to overcome issues inhibiting today's PINNs.

\begin{figure*}[htbp]
\centering
\includegraphics[width=1.0\linewidth]{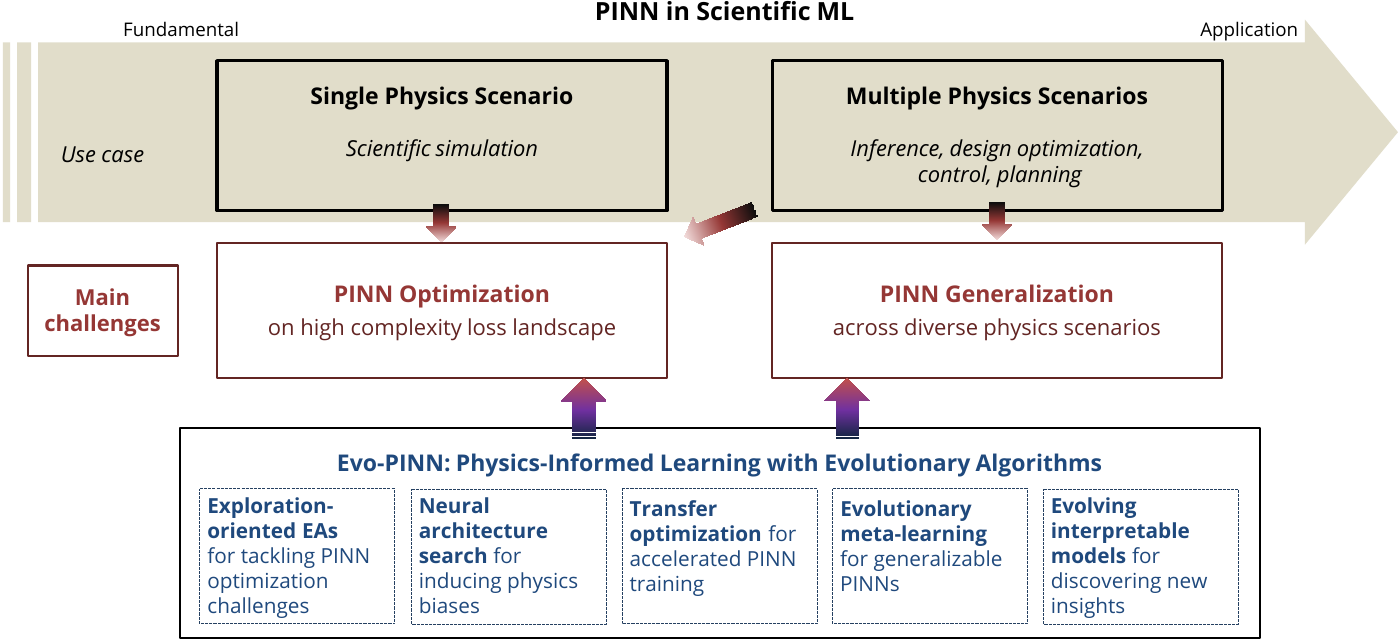}
\caption{Boosting scientific ML with evolutionary optimization of physics-informed neural networks (Evo-PINN).}
\label{fig:pine-summary}
\end{figure*}

In light of the above, this paper examines PINNs through the lens of their optimization and generalization challenges. As evidenced by Fig.~\ref{fig:pinn-popularity}, there is a vast amount of innovative PINN literature, however, this paper specifically focuses on a curated subset that best illustrates both these optimization and generalization challenges, and the exciting opportunities for novel, targeted algorithms and methods. Notably, most PINN implementations today make use of gradient-based optimization algorithms. The ubiquity of the method may be attributed to inertia from the primacy of stochastic gradient descent (SGD) in conventional data-rich deep learning tasks. However, SGD is no silver bullet. Relative to standard deep learning, physics-informed loss functions are not only highly non-convex but also exhibit disparities in the magnitudes of back-propagated gradient signals between the data and physics loss terms (see \ref{sec:pinn-landscape}). These issues lead to pathologies that hinder effective training \cite{bischof2021multi,wang2021understanding}. The resulting optimization trajectory is littered with spurious minima, making it extremely challenging to find globally optimum solutions that satisfy physics. There is a high chance for gradient descent to get trapped in a local minimum that fails to satisfy the requisite physics, a possibility discussed in greater detail in Section~\ref{sec:pinn-challenges}.

%In order to fulfill the vision of data-efficient scientific machine learning, this paper asserts the need for algorithmic advances \textit{beyond gradient descent} to move the needle for PINNs. 
Fig.~\ref{fig:pine-summary} provides a schematic summary of some of the main computational challenges inhibiting the vision of physics-informed machine learning towards AI for Science, and some corresponding proposed solutions. In Section~\ref{sec:pine-ideas}, arguments are presented in support of our main claim: that algorithmic ideas developed over the years within the computational intelligence community may be uniquely well-suited for physics-informed learning. Of particular interest are the gradient-free evolutionary algorithms (EAs) that suffer less from spurious local minima due to their exploration-oriented global search capacity \cite{wong2021can, gupta2024neuroevolving}. An associated research direction is the synergy of EAs with gradient descent (in the spirit of memetic algorithms), thereby marrying the best of both worlds. While population-based EAs may be advantageous in navigating highly non-convex loss landscapes and balancing trade-offs between conflicting objectives, local search via gradient descent can facilitate rapid convergence to precise solutions.

In addition to optimizing network parameters, EAs enable the discovery of bespoke architectures for physics-informed learning \cite{kaplarevic2023identifying}. Evolutionary mechanisms could also prove useful for guiding training data generation in areas with higher errors on governing physics equations, having shown evidence of improved convergence to correct solutions~\cite{daw2023mitigating}. Accurate solutions to novel PDEs can be rapidly obtained through evolutionary transfer optimization, leveraging multiple knowledge sources during optimization for specific target tasks~\cite{wong2021can}. Another exciting track is to cast EAs as a meta-learner of generalizable PINN models~\cite{wong2024generalizable}. This may be achieved through a bilevel formulation where an upper-level EA searches for PINN configurations that can be quickly fine-tuned for various downstream physics tasks at the lower-level. 

The advancement of Evo-PINNs is expected to have far-reaching impact on establishing PINN methodologies at the forefront of scientific machine learning. However, there are only a handful of related works in the literature, implying huge scope for future research and development. %Yet, surprisingly  that explore such ideas

\section{Overview of PINN Methodologies}

PINNs are designed to learn a mapping between the input and output variables while satisfying specified mathematical constraints that represent the physical phenomenon or dynamical process of interest~\cite{raissi2019physics}. Given their ubiquity in modern science and engineering, PDEs are one of the most studied physics priors integrated in PINNs. Without loss of generality, we consider a physics-informed learning problem with input variables $x$ (spatial dimension) and $t$ (time dimension), and an output variable $u$ satisfying PDEs of the general form: % of interest 
\begin{subequations} \label{eq:pde_ibc_eqn}
    \begin{align}
        & \text{PDE:} & \mathcal{N}_\vartheta[u(x,t)] &= h(x,t), & x\in\Omega, t\in(0,T] \label{eq:pde_eqn} \\
        & \text{IC:} & u(x,t=0) &= u_0(x), & x\in\Omega \label{eq:ic_eqn} \\
        & \text{BC:} & \mathcal{B}[u(x,t)] &= g(x,t), & x\in\partial\Omega, t\in(0,T] \label{eq:bc_eqn}
    \end{align}
\end{subequations} 
where the general differential operator $\mathcal{N}_\vartheta[u(x,t)]$ contains PDE parameters $\vartheta$ and can include linear and/or nonlinear combinations of temporal and spatial derivatives of $u$, and $h(x,t)$ is an arbitrary source term in the domain $x\in\Omega, t\in(0,T]$. Equation~\ref{eq:ic_eqn} specifies the initial condition (IC), $u_0(x)$, at time $t$=$0$, and Equation~\ref{eq:bc_eqn} specifies the boundary condition (BC) such that $\mathcal{B}[u(x,t)]$ equates to $g(x,t)$ at the domain boundary $\partial\Omega$, where $\mathcal{B}[\cdot]$ can either be an identity operator (Dirichlet BC), a differential operator (Neumann BC), or a mixed identity-differential operator (Robin BC).

\begin{figure*}[htbp]
\centering
\includegraphics[width=1.0\linewidth]{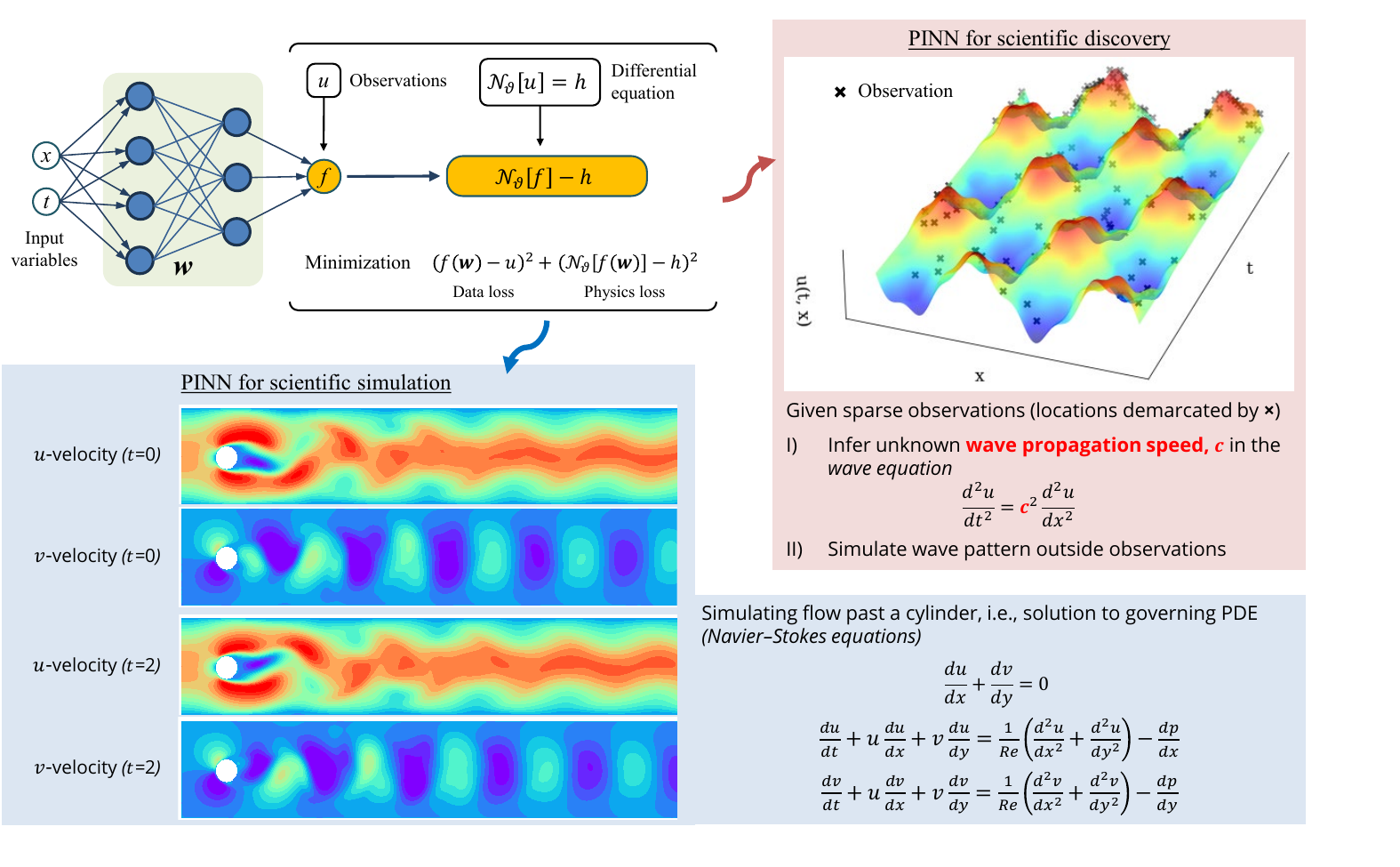}
\caption{PINN handles both scientific simulation and scientific discovery. Example applications are: 1) accurate simulation of transient vortex shedding behind a cylinder obtained by learning a PINN for a single problem solely using the governing physics (2D transient N-S equations); and 2) inference of the wave propagation speed, $c$, given sparse observations (locations demarcated by $\boldsymbol{\times}$), by learning an inference PINN model with a partially specified 1D transient wave equation.}
\label{fig:pinn-schemetic}
\end{figure*}

\subsection{Physics-Informed Loss} \label{sec:pinn-loss}

PINNs are distinct from DNNs in that additional physics priors are incorporated as an essential component of its physics-informed loss function:
\begin{equation} \label{eq:loss_fn}
    \mathcal{L}_{\text{PINN}} = \lambda_{physics}\ \mathcal{L}_{physics} + \lambda_{data}\ \mathcal{L}_{data},
\end{equation}
where the relative weights $\lambda_{physics}$ and $\lambda_{data}$ control the component-wise trade-off. The physics-informed component can consist of multiple PDE, IC, and BC loss terms: % between the two components
\begin{subequations} \label{eq:physics_loss_fn}
    \begin{align}
    \mathcal{L}_{physics} &= \lambda_{pde}\ \mathcal{L}_{pde} + \lambda_{ic}\ \mathcal{L}_{ic} + \lambda_{bc}\ \mathcal{L}_{bc} \\
    \mathcal{L}_{pde} &= \lVert \mathcal{N}_\vartheta[f(\cdot;\boldsymbol{w})] - h(\cdot) \rVert _{L^2(\Omega \times (0,T])}^2 \\
    \mathcal{L}_{ic} &= \lVert f(\cdot,t=0;\boldsymbol{w}) - u_0(\cdot) \rVert _{L^2(\Omega)}^2 \\
    \mathcal{L}_{bc} &= \lVert \mathcal{B}[f(\cdot;\boldsymbol{w})] - g(\cdot) \rVert _{L^2(\partial\Omega \times (0,T])}^2
    \end{align}
\end{subequations} 
where $f(\cdot;\boldsymbol{w})$ is the PINN output function with network parameters $\boldsymbol{w}$, evaluated on a set of label-free training points (collocation points) sampled from the respective spatio-temporal domain. An illustrative schematic of a PINN model is presented in Fig.~\ref{fig:pinn-schemetic}. Evaluation of differential operators, e.g., $\frac{d\ f(\cdot;\boldsymbol{w})}{dt}$, $\frac{d\ f(\cdot;\boldsymbol{w})}{dx}$, and $\frac{d^2f(\cdot;\boldsymbol{w})}{dx^2}$, are generally required for computation of the physics-informed loss on these training samples. They can be conveniently computed by automatic differentiation~\cite{raissi2019physics} and/or numerical differentiation~\cite{chiu2022can,wong2023lsa} methods at any sample location.

Fundamentally, PINNs arrive at an accurate and physics-compliant prediction for a desired physics scenario by forcing its outputs $f(\cdot;\boldsymbol{w})$ to satisfy Equation~\ref{eq:pde_ibc_eqn} through minimizing the physics-informed loss (Equation~\ref{eq:physics_loss_fn}). The effectiveness of the optimization of network parameters $\boldsymbol{w}$ through the physics-informed loss is a primary determinant of the accuracy of the PINN model. In contrast, the typical data-driven loss (as common in a DNN) computes the MSE between the model prediction $f(x_i, t_i; \boldsymbol{w})$ and available target $u_i^{label}$ over $i$=$1,...,n$ labeled samples:
\begin{equation} 
    \mathcal{L}_{\text{DNN}} = \mathcal{L}_{data} = \frac{1}{n}\sum_{i=1}^{n} \left( u_i^{label} - f(x_i, t_i; \boldsymbol{w}) \right)^2 \label{eq:dnnloss}
\end{equation}

\begin{figure}[htbp]
\centering
\includegraphics[width=1.0\linewidth]{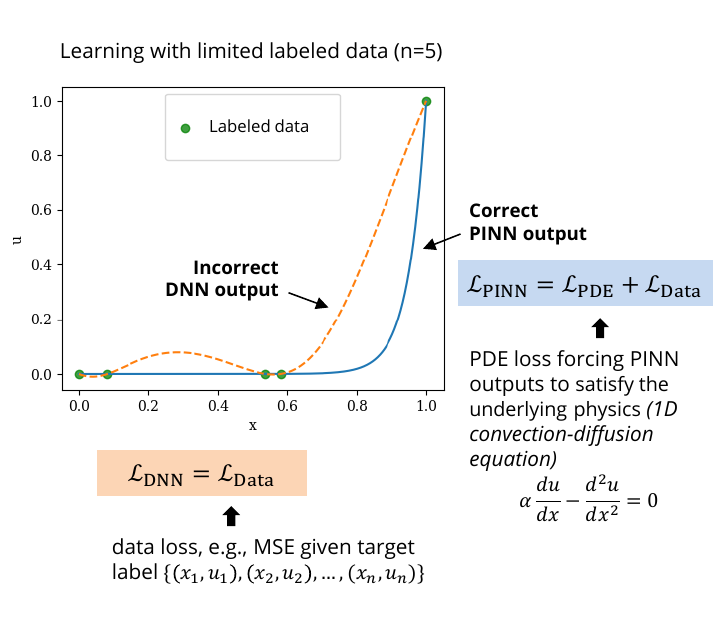}
\caption{Illustrative example showing how a PINN model can make accurate prediction by learning with physics laws (1D convection-diffusion equation) and few labeled data, whereas a conventional DNN model fails to learn the correct output.}
\label{fig:pinn-data-free}
\end{figure}

\subsection{Physics-Informed Learning Advantages} \label{sec:pinn-advantages}

The advantages of physics-informed learning are demonstrated in Fig.~\ref{fig:pinn-data-free}. Given only a few labeled data ($n$=$5$), a DNN model (with sufficient representation capacity) remains unable to correctly reconstruct the entire system, although it fits all training samples well. In contrast, a PINN model can accurately learn the solution from the same set of labeled data. By enforcing physics constraints on a set of \textit{label-free} training samples ($n_{pde}$=$1000$) uniformly drawn from the computation domain utilizing the aforementioned PINN loss, a PINN can learn the solution even without any labeled data. The governing physics in this example is the convection-diffusion equation $\alpha \frac{du}{dx} - \frac{d^2u}{dx^2} = 0$ with PDE parameter $\alpha$, a ubiquitous physics model that describes the distribution of a scalar quantity (e.g., mass, energy, temperature) in the presence of convective transport and diffusion~\cite{stynes2005steady}. %in science and engineering 

\subsection{PINN for Scientific Simulation}  \label{sec:pinn-forward}

PDEs offer concise mathematical descriptions of diverse physics phenomena in nature. With PINNs, it is possible to simulate an entire physical system by training the model to approximate the solution of the PDEs specific to the target physics, without any labeled data (i.e., solely using the physics-informed loss). A good example of this is the fluid dynamic simulation of vortex shedding behind a cylinder depicted in Fig.~\ref{fig:pinn-schemetic}, which was obtained by using a PINN to solve the transient 2D Navier-Stokes (N-S) equations~\cite{temam2024navier}.

Compared to classical numerical solvers employing the finite difference, finite element, or finite volume method~\cite{peiro2005finite}, PINNs have the significant advantage of being mesh-free. Meshing, which is the process of domain discretization, is an essential, yet non-trivial task in classical numerical methods. Numerical methods can fail to find the correct solution if the meshing is not appropriately done. On the other hand, PINNs can provide a solution without similar mesh requirements. Additionally, the solution obtained through PINN is differentiable, and its compact representation requires lower memory demand for storage. This makes PINNs a promising alternative to traditional numerical methods for simulating physics phenomena; an extensive review is presented in~\cite{cuomo2022scientific}. 

While most PINN studies address single physics problem, a PINN model can also incorporate additional input parameters, such as geometry, Reynolds number, or material properties~\cite{wandel2020learning, zhang2020physics, mahmoudabadbozchelou2021rheology}. These models, also referred to as parameterized PINN, can then provide predictions across a set of physics scenarios, rather than just one.\\

\subsection{PINN for Scientific Discovery} 

In addition to directly simulating physical behavior, PINNs can be seamlessly extended to various hypothesis inference problems common in scientific discovery. These problems typically involve an incomplete and potentially noisy set of observed data points, and the goal is to infer information about the system through optimizing the hypothesis space. The loss function used consists of both physics-informed (to capture relevant physics) and data (observations) components (Equation~\ref{eq:loss_fn}). The hypothesis spaces of the inference problem are included as optimization variables along with the network parameters during training; see example in Fig.~\ref{fig:pinn-schemetic}. %to train PINN models for inference 

The PINN models are uniquely suited for hypothesis inference when compared to numerical solvers and can effectively synergize theory and observation data from various sources such as sensors and imaging, even for ill-posed PDE problems. It can be combined with probabilistic inference methods (e.g., Bayesian-PINN~\cite{yang2021b}), with the probabilistic framework offering a mathematical tool to handle uncertainties arising from both noisy data and physics assumptions~\cite{yang2021b, viana2021survey}.

In current literature, PINN models have been applied to: \textit{1.) Inference of system-level parameters} such as the underlying governing equations, constitutive relationships or system parameters (e.g., Reynolds number, wave speed) based on observations~\cite{tartakovsky2020physics,chen2021physics,kharazmi2021identifiability,shi2021physics,oszkinat2022uncertainty}. \textit{2.) Inference of unobserved field variables} across the entire domain based on limited observations, e.g., predicting pressure from velocity in particle imaging velocimetry~\cite{raissi2020hidden}. Many have employed PINNs towards the inference of domain field variables in various problems spanning fluid dynamics~\cite{zhu2021machine}, heat transfer~\cite{liu2022temperature}, and biomedical sciences~\cite{cai2021artificial,ruiz2022physics}. \textit{3.) PDE-constrained design optimization} towards a set of desired (observation) outcomes. The PINN is used as an adjoint optimizer, enabling the design parameters to be optimized towards the desired outcome while fulfilling all specified relevant physics during training. This has been successfully demonstrated on topology optimization~\cite{lu2021physics} and metamaterial~\cite{fang2019deep} design problems. % experiment with only velocity observations

\begin{figure*}[htbp]
\centering
\includegraphics[width=1.0\linewidth]{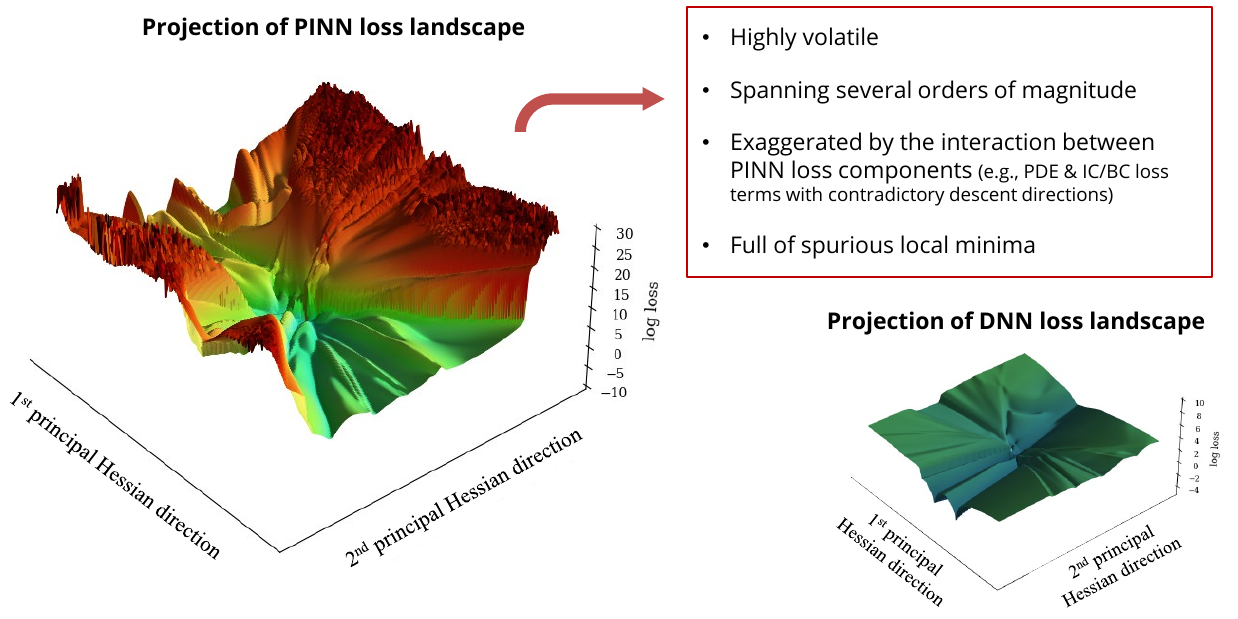}
\caption{The plots contrast the local loss landscapes of PINN and DNN, for the convection-diffusion problem presented in Subsection~\ref{sec:pinn-advantages}. After 100,000 training iterations, we project their corresponding loss values onto the plane spanned by the first two principal Hessian directions as per previously reported literature~\cite{krishnapriyan2021characterizing} to get an intuition for the loss landscape.}
\label{fig:pinn-landscape}
\end{figure*}

\section{Optimization and Generalization Challenges} \label{sec:pinn-challenges}

However, PINN's advantages come at the cost of increased difficulty in optimizing PINN model parameters, with many early papers showing long training times and limited predictive performance on more complex problems~\cite{raissi2019physics, fuks2020limitations, ji2021stiff, krishnapriyan2021characterizing}. Unsuccessful PINN training (premature convergence) can lead to unacceptable (unphysical) simulation outcomes. Moreover, parameterized PINNs can still face issues with generalizability since the accuracy of interpolation-based predictions on a new problem is not guaranteed, and training a well-generalizable PINN model may not be an easy task. In this section, we shed light on some contributing factors to the optimization and generalization challenges---two pre-dominant hurdles to the wide-spread adoption of PINNs in real-world applications today---and how they may originate from fundamental limitations in gradient-based approaches.

\subsection{Challenges in Optimizing PINN Models}

In many studies involving PINN, it was observed that the training iterations required for convergence far exceed those needed for typical data-driven DNN. We relate the optimization difficulty of PINNs to a few factors below. \\

\subsubsection{Complex Loss Landscape} \label{sec:pinn-landscape}

The physics-informed loss usually consists of multiple PDE and IC/BC loss terms (Equation~\ref{eq:physics_loss_fn}), each with distinct scales and characteristics during PINN training. For instance, the scale of PDE loss can span several orders of magnitude depending on the underlying differential operators and the complexity of the solution, and this range is only revealed during training. Additionally, some loss terms change more dramatically than others during training, as shown by prior analyses through the lens of neural tangent kernel (NTK)~\cite{wang2021eigenvector,wang2022and}. These characteristics make the proper scaling of different loss terms challenging.

In addition, the PINN optimization landscapes are littered with spurious minima and are more complex than those of usual data-driven DNN models. Fig.~\ref{fig:pinn-landscape} illustrates the loss landscapes of PINN and DNN models for the convection-diffusion problem presented in Subsection~\ref{sec:pinn-advantages}. To investigate the topological differences in their loss landscapes, we first computed the Hessians of their respective loss functions (i.e., $\mathcal{L}_{\text{PINN}}$ or $\mathcal{L}_{\text{DNN}}$) with respect to the model parameters: $\mathbf{H} = \nabla^2\ \mathcal{L}(\boldsymbol{\boldsymbol{w}})$. We then extracted the leading two eigenvectors and eigenvalues of the Hessian (i.e., first two principal Hessian directions) which correspond to the directions of highest curvature in the loss landscape. By perturbing the trained models along these dominant eigenvectors and evaluating the resulting loss values, we can obtain a more informative characterization of the local topology than perturbing the model parameters in random directions~\cite{krishnapriyan2021characterizing}. This visualization provides evidence of a much higher degree of complexity and ruggedness of the loss landscape for PINN compared to usual data-driven DNN models. The smoothness of the latter is conducive to gradient-based algorithms that can easily find descent paths from the point of initialization to a good local optimum that fits the training data well. This outcome corroborates the primacy of SGD and its variants in today's deep learning tasks. In contrast, the contradictory descent directions from different loss terms encountered during PINN training can adversely impact optimization~\cite{elhamod2020cophy,xiang2022self}. It is worth noting that when the loss terms are purely physics-based (e.g., PDE and IC/BC), they are not theoretically contradictory. However, during the optimization process, especially when the solution is still far from the true optimum, the local gradient directions of these loss components can still become practically conflicting. This issue arises due to the combined effects of neural network expressivity, the structure of the loss landscape (e.g., scales and convergence dynamics), and the limitations of optimizers, which can cause certain objectives to dominate or suppress others during early or intermediate stages of training.

Although these powerful gradient-based algorithms are inseparable from the recent success of deep learning, they are no silver bullet. For instance, gradient-based methods are susceptible to being trapped in spurious local minima, which is often the case in the context of PINN optimization. Typical deep learning applications do not require a fully converged solution, as they prioritize generalization performance over training precision. However, this is not the case for PINNs, where reaching the global minimum is essential to ensure the model's fidelity to the underlying physics. A small violation of the PDE constraints can result in an unphysical PINN solution while a small violation of the BC or IC constraints corresponds to an entirely different PDE problem~\cite{wang20222,wang2024respecting}.\\ %This issue is especially detrimental in time-varying dynamical problems where small violations early in time can cause errors to propagate and accumulate in later times.

\subsubsection{Initial Local Minimum Trap and Spectral Bias}

PINN models employing standard activation functions (such as \textit{tanh}, \textit{sigmoid}) when initialized using common techniques (such as the Xavier method) tend to produce flat output functions at initialization, making them immediately susceptible to a local minimum trap whereby the PINN loss already sees a minimized PDE loss term. This theoretical insight, discussed in~\cite{wong2022learning,gupta2024multi,jin2024fourier}, applies across a range of physics scenarios where the underlying PDE is trivially satisfied by  flat functions. A strong bias around flat functions can therefore be difficult for gradient descent to escape, especially when the true solution, that must simultaneously minimize \emph{all} PDE and IC/BC loss terms, is not accessible near the point of initialization.

Another closely related issue stems from the known saturation property of activation functions commonly used for PINN. Such saturation property can cause backpropagated gradients to vanish whenever the network weights are shifted towards a larger distribution, as driven by the need to approximate higher frequencies and steep gradients common in physics. This makes further gradient descent of the network weights difficult, adversely causing a learning bias towards smooth functions (lower frequencies are learned better than higher frequencies). This known tendency---also called \textit{spectral bias}---has been studied through the lens of NTK~\cite{rahaman2019spectral,xu2019training}, revealing a significant discrepancy in convergence rates for different frequency components in the output during training. While this has been regarded as a natural regularizer with potential benefits to the generalization performance of DNNs, the consequences are highly deleterious for PINNs. Spectral bias makes it difficult for PINNs to accurately learn complex nonlinear physics behaviors~\cite{bonfanti2024challenges}, e.g., in the vicinity of steep gradients characterizing shock waves. These local discrepancies further deter propagation of the correct initial and boundary conditions into the computation domain, which has been suggested to be critical for successful PINN training~\cite{wang2024respecting,daw2022rethinking}.

This saturation also translates to increased difficulties for gradient-based algorithms to alter the behavior of PINN output far away from the input origin. As a result, PINN struggles to learn over long time horizons, partly because the learning bias towards smooth functions can cause the gradient descent to prematurely converge to a trivial solution (satisfying the PDE) before the initial and boundary conditions can properly propagate to a later time~\cite{wang2023long}. An associated tendency for gradient-based PINN training to be drawn towards unstable fixed point solutions of dynamical systems was analyzed in \cite{rohrhofer2023role}, with experiments suggesting that they form strong basins of attraction in the PINN loss landscape.\\

\subsubsection{Sensitivity to Sample Size and Location}

The training samples of the PINN model are typically uniformly drawn from the input domain and are equally important everywhere due to their physical correspondence with the spatio-temporal extent of the problem. While the use of mini-batch samples is a standard procedure in gradient-based training, it remains an open question how one can efficiently reduce the batch size (and hence the training cost) while maintaining the desired PINN model accuracy~\cite{lau2024pinnacle}. Using a large batch size is crucial for the accuracy of PINNs due to the potential need to learn high frequencies or complex structures in the \textit{a priori} unknown target function \cite{ooi2024importance}. However, this inevitably increases training costs. Furthermore, oversampling on a high loss region may adversely slow convergence since it can be harder to learn such local complex gradient behaviour---a cause of stiffness during training~\cite{wong2023lsa,fuks2020limitations,ji2021stiff}. On the other hand, an overly small batch size will result in noisy (big jump in loss values) and inconsistent local gradient signals~\cite{wu2024ropinn}, thereby causing issues during gradient descent.\\

\begin{figure*}[htbp]
\centering
\includegraphics[width=1.0\linewidth]{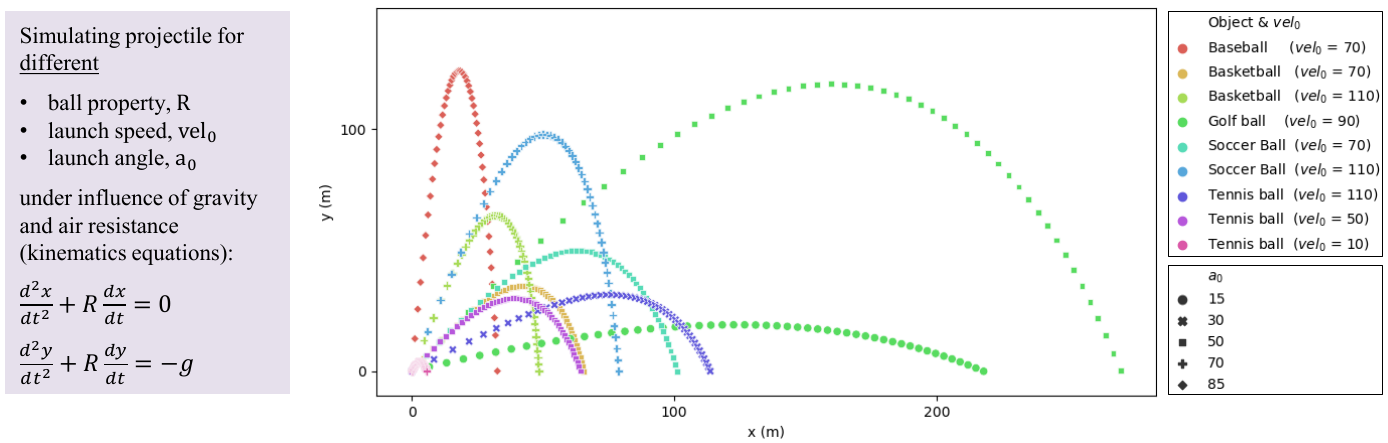}
\caption{Example of learning PINN for simulating a distribution of tasks with \textit{a priori} unknown input conditions in a what-if analysis: projectiles of ball with different physics properties and initial conditions (launch angle and launch speed).}
\label{fig:task-distribution}
\end{figure*}

\subsubsection{Current Research Directions}

Since the seminal work of Raissi \textit{et al.}~\cite{raissi2019physics}, architecture design and training methods for PINNs have continued to evolve. Recent advancements often combine multiple learning strategies to tackle the aforementioned PINN optimization issues~\cite{wang2023expert,wang2024piratenets,wang2025gradient}. \textit{Curriculum learning} improves convergence by initiating training on a better-conditioned, simpler version of the original problem and then systematically advancing to target problem~\cite{krishnapriyan2021characterizing}. Physics-informed learning often succumbs to premature convergence for challenging problems, however, the same PDE equation with altered parameters or IC/BCs may be a simpler problem. In this scenario, one can start training the PINN model on the easier problem and then gradually vary the problem during training to approach the target problem. This approach assumes that each training step brings the model closer to the optimal solution of the next (harder) problem, thereby improving the next round of training. Hence, curriculum learning can be viewed as an initialization process to guide the model toward the desired solution utilizing \textit{a priori} understanding of how the system behaves as its problem parameters change. \textit{Adaptive loss-balancing strategy} can also be considered as a form of curriculum learning, where the training is initiated with the loss weights corresponding to a simpler optimization problem. It involves assigning appropriate weights to balance different terms in the PINN loss functions~\cite{xiang2022self, wang2022and, wang2021understanding, wang2021eigenvector, elhamod2020cophy}, and is regarded as an important technique to improve convergence. %after each transition step, providing a better starting point for

Recent studies have proposed enhancements to PINN learning by leveraging network architectures that incorporate Fourier feature embeddings, skip connections, and random weight factorization~\cite{wang2023expert,wang2024piratenets,wang2025gradient}. PINNs with skip connections and periodic activation functions retain a greater proportion of gradients during backpropagation from output to input, thereby mitigating spectral bias caused by activation saturation~\cite{sitzmann2020implicit, wong2022learning, jin2024fourier}. Adaptive techniques that adjust the slope of activation functions have also been introduced to further control saturation and enhance training efficiency~\cite{jagtap2020adaptive, zeng2022adaptive}. These methods embed inductive biases---such as high-frequency behavior characteristic of certain PDEs---directly into the network architecture. Additionally, physics-informed cell representations have been developed to address spectral bias away from the input origin by disentangling network parameters from the input coordinates~\cite{kang2023pixel}. Convolutional architectures also achieve this parameter-input decoupling, allowing for larger, more expressive networks that scale effectively to complex, large-scale problems~\cite{wandel2020learning}. %enable PINNs to  via trainable multiplicative factors 

Some studies have proposed using numerical differentiation (ND)-based methods for the differential operators in the physics-informed loss, with ND being potentially cheaper to compute relative to automatic differentiation~\cite{chiu2022can, wong2023lsa}. More importantly, ND is more robust to sample density and potentially advantageous in the context of reducing sample requirement in establishing a strong correlation between physics-informed loss and model accuracy. This is because ND methods, by definition, use Taylor series expansion (TSE) to construct a \emph{low-order local approximation} for the derivative terms. They can provide structural biases and discourage extreme gradients to reduce overfitting to better learn the approximate solution~\cite{chiu2022can}. They can also connect neighboring samples and facilitate the propagation of information from the IC/BC to interior sample points, thereby reducing any tendency to overfit especially near boundaries~\citep{wong2023lsa}.

\subsection{Challenges in Learning Generalizable PINN Model} \label{sec:pinn-generalize}

Given both simulation (hypothesis evaluation) and inference capabilities, PINNs are versatile and can be applied to various applications common in science such as model-predictive-control, scenario planning, and design exploration~\cite{arnold2021state, gokhale2022physics, gungordu2022robust, mowlavi2023optimal}; see Fig.~\ref{fig:task-distribution}. The commonality to these applications is the need for multiple accurate and physics-compliant predictions across a series of related problems. These repeated evaluations usually differ by modifications to the physics scenario such as changes in geometric shape, PDE parameter, IC, and BC.\\ %the example in 

\textit{1) How Generalizable Is A PINN Model?} Most parameterized PINN models (ref. to Subsection~\ref{sec:pinn-forward}; they are designed to learn a parametric family of physics scenarios) make predictions on new scenarios unseen during training through interpolation. This interpolation is usually physics-agnostic, with similar negative implications for generalizability as predictions from typical DNNs~\cite{wong2024generalizable}. Consequently, the effectiveness of parameterized PINNs depends significantly on problem complexity. Given the difficulties in optimizing a PINN model for a single scenario, the cost of performing physics-informed learning across a set of scenarios can be prohibitive~\cite{iwata2023meta}. Moreover, the problem's parameterization needs to be determined at the learning stage, which reduces the flexibility of the parameterized PINN model when applied to new physics scenarios.\\ %Contrary to its promise, the

\textit{2) How to Train A Generalizable PINN Model in The Data-Scarce Regime?} Meta-learning approaches provide a systematic framework for discovering good weight initializations that can be quickly adapted---\textit{with minimal training}---to  new scenarios (tasks) in a problem distribution. Hence, meta-learning of a generic PINN model initialization can similarly accelerate physically precise predictions across diverse, multi-modal problem distributions, and holds immense promise for addressing generalization challenge of PINN~\cite{de2021hyperpinn,huang2022meta, toloubidokhti2023dats,liu2022novel,iwata2023meta,penwarden2023metalearning,cho2024hypernetwork,chen2024gpt,cho2024parameterized}.

There are still a number of unresolved issues in meta-learning of a generalizable PINN model. Recent PINN studies showed inferior results for model-agnostic meta-learning (MAML), potentially due to insufficient task-specific adaptation and non-convergence in the meta-learning algorithm's outer loop optimization~\cite{penwarden2023metalearning, cho2024hypernetwork}.  A fundamental assumption of popular gradient-based meta-learning algorithms---such as MAML~\cite{finn2017model} and its variant, Reptile~\cite{nichol2018reptile}---is that the models can learn well on each task after a few gradient descent updates from the initialization. However, this is not the case for PINNs because physics-informed losses are much more challenging to optimize and require extensive training~\cite{iwata2023meta}. With only a handful of task-specific gradient descent updates, gradient-based meta-learning algorithms may remain susceptible to local minima from noisy or even deceptive gradients. %-specific training .  to achieve good convergence rone to getting stuck in  from a batch of diverse training tasks can be noisy or even deceptive As a result, r

\section{Evo-PINN: Physics-Informed Learning with \\Evolutionary Algorithms} \label{sec:pine-ideas}

Despite their theoretical advantage in diverse scientific domains, Section~\ref{sec:pinn-challenges} has highlighted several practical challenges in today's PINN learning and optimization. The cost of optimization, navigating spectral bias and local minima traps, has even seeded a sense of pessimism among computational scientists about the utility of these techniques over standard numerical methods \cite{mcgreivy2024weak, grossmann2024can}. In this section, we present our hypothesis that many algorithmic ideas developed in the computational intelligence community may in fact be uniquely well positioned to tackle the issues plaguing PINNs. One exciting research direction is to achieve physics-informed learning by synergizing evolutionary algorithms with local gradient descent signals, thereby leveraging the best of both worlds to quickly arrive at physically accurate and generalizable PINN models. In what follows, we examine this and other promising algorithmic directions that we believe to warrant sustained investigation. These directions highlight key opportunities for the evolutionary optimization of PINNs (Evo-PINNs) to advance scientific machine learning. 

\subsection{Exploration-Oriented Evolutionary Algorithms for Tackling PINN Optimization Challenges}

\subsubsection{Neuroevolution Can Avoid Local Minima Traps}

In looking at a PINN's unique optimization challenges, the need for new algorithms to overcome local minima traps caused by flat outputs at initialization, or by the attractiveness of unstable fixed point solutions to dynamical systems, becomes apparent. Gradient descent is a point-based optimization method that is highly sensitive to the issues of local minima or learning bias magnified by PINN loss functions. In contrast, in silico evolutionary computation offers a class of highly parallelizable approaches for guiding \textit{populations} of candidate solutions towards globally-optimal regions of a search space via randomized mechanisms inspired by biological evolution~\cite{miikkulainen2021biological}. We broadly refer to any such algorithmic realization as an \textit{evolutionary algorithm} (EA) and its application to the training of neural networks as \textit{neuroevolution}.

The operators of EAs are typically based on principles of natural selection. They do not require derivative computations of the loss function to be optimized and are therefore gradient-free. Importantly, as population-based methods, the implicit parallelism of EAs implies that they have lesser tendency to get trapped in a single local minimum~\cite{back1996evolutionary,gupta2024neuroevolving}. What is more, since their search is typically invariant to order-preserving transformations of the loss function~\cite{ollivier2017information}, EAs are less susceptible to flat output functions with small (vanishing) gradients. For these reasons, they are deemed as potential alternatives to SGD, particularly in the context of scientific machine learning where the goal is to converge to physically precise solutions represented by global optima with zero PINN loss~\cite{wong2021can}. In addition to its global search capacity, an evolved population naturally provides an ensemble of neural networks that affords the possibility of accurate predictions with robust uncertainty quantification---as was shown through the application of a swarm-based evolutionary procedure to PINN optimization~\cite{davi2022pso,davi2023multi}. The ability to provide uncertainty quantification and probabilistic forecasts is crucial for scientific modelling and control problems \cite{fasel2022ensemble}.\\

\subsubsection{Pareto Optimization Balances Physics and Data Losses}
\label{sec:pareto-optimization}

The existence of multiple loss terms---see Eqs. (\ref{eq:loss_fn}) and (\ref{eq:physics_loss_fn})---makes PINN training intrinsically a multi-objective optimization problem. Since the terms $\mathcal{L}_{data}, \mathcal{L}_{pde}, \mathcal{L}_{ic}$, and $\mathcal{L}_{bc}$ represent different physical quantities, their magnitudes may be incommensurable. In this case, the role of the scalarizing weights is not only to express preferences between the losses, but also to compensate for differences in their magnitudes. While a perfect physics solution would, in theory, simultaneously satisfy all PINN loss terms, the expressive capacity of a neural network is always limited by its (necessarily) finite size in practice. As argued in \cite{rohrhofer2023role,choong2023jack}, capacity limitations naturally give rise to conflicts and trade-offs between loss terms. This, in conjunction with the extreme ruggedness of the physics-informed learning objective, implies that a neural network, despite being a theoretical universal function approximator, is not guaranteed to find a solution that simultaneously minimizes all the terms \cite{bischof2021multi}. When data-driven terms are also included, such as observational data or empirical measurements, true contradictions may arise as the data may be noisy, incomplete, or governed by partially unknown physics that differs from the modeled equations. In such cases, the multi-objective optimization must reconcile inherently conflicting objectives between data fitting and physics adherence. Identifying an appropriate aggregation of objectives to arrive at good (physically satisfactory) trade-off solution is also non-trivial \cite{wei2023select} Moreover, the shape of the Pareto front formed by the multiple objectives is not known beforehand \cite{rohrhofer2023apparent}, with simple weighted-sum aggregations being insufficient to support all parts of non-convex fronts \cite{heldmann2023pinn}.

Multi-objective EAs (MOEAs) are tailor-made to address these issues  \cite{coello2007evolutionary}. In particular, Pareto dominance-based variants of MOEAs~\cite{deb2023generalized} do not require the multiple loss terms to be scalarized, and hence are less sensitive to incommensurable objectives. Their population-based search mechanisms also allow for free-form convergence towards the entire Pareto front in a single run of the algorithm, irrespective of the front's shape. These salient features have naturally begun to attract the attention of PINN researchers, although associated developments are still at a nascent stage. In \cite{lu2023nsga}, the non-dominated sorting genetic algorithm \cite{deb2002fast} was hybridized with a variant of SGD (namely, Adam \cite{kingma2014adam}) for training the parameters of a PINN. The exploration capacity of the EA was shown to help overcome local minima and enable precise satisfaction of the initial and boundary conditions encoded in the physics-informed loss. Given the computational cost of a hybrid framework, a pure population-based multi-objective search algorithm was studied for PINN training in \cite{davi2023multi}, leading to calibrated uncertainty quantification by examining the empirical distribution of the ensemble's predictions.

In addition to using MOEAs for PINN parameter training, there have been efforts towards evolving the model's hyperparameters (including the scalarizing weights and activation functions). The method yields a Pareto set of hyperparameters where each setting offers an optimal trade-off between the data and physics losses. Practitioners can then decide post-hoc which setting they wish to pursue \cite{de2022mopinns}. An analysis of such post-hoc decision-making reveals that solutions close to the convex bulge of the Pareto front typically lead to performant PINNs \cite{wei2023select}. Evolutionary algorithmists will immediately notice a connection between this observation and the literature on knee-point driven MOEAs \cite{yu2022survey}, which could serve as a valuable resource for hyperparameter tuning in PINNs. \\

\subsubsection{Memetic Algorithms Can Navigate Complex PINN Landscapes}

There is a clear theoretical rationale supporting Evo-PINNs, yet there are only a handful of related works in the literature. A barrier to the widespread use of EAs has been their high sample complexity, a consequence of updating and evaluating populations of solutions over multiple iterations. It is useful to consider hybridizing global evolutionary search with local gradient signals---thus inheriting the merits of both approaches. While the former can serve as an efficient basin-hopper for exploring rugged PINN landscapes, the latter facilitates rapid convergence to stationary points by exploiting local derivative information of loss functions. % that explore such ideas

There exists a long history of research in evolutionary computation studying synergies of this kind, with adaptive memetic algorithms forming a class of methods tailored for this purpose~\cite{ong2006classification}. In the deep learning literature, gradient-guided evolutionary approaches have been designed for training neural networks~\cite{yang2021gradient,xue2022ensemble}, with Cui \textit{et al.}~\cite{cui2018evolutionary} proving a simple theorem that the performance of an evolutionary stochastic gradient descent method with elitist selection strategy never degrades with increasing iterations. However, to the best of our knowledge, there is no comparable body of work for PINN. Davi and Braga-Neto~\cite{davi2022pso} demonstrated that combining a swarm optimization algorithm (a population-based method like EAs) with gradient descent could produce an ensemble of PINN models that out-performed pure gradient-based learning. This result illuminates unexplored avenues for unifying SGD with gradient-free EAs as complementary algorithms within a single framework for finding globally optimum PINNs.

The crafting of synergistic optimization frameworks that meld gradient-based and gradient-free methods, however, remains challenging—particularly in identifying interaction mechanisms that effectively harness the strengths of both approaches. Key questions to address include balancing the intensity of SGD versus EAs, ascertaining when gradient descent should take over from evolution, or selecting an appropriate subset of the evolving population for fine-tuning with gradient descent (as doing so for the entire population would be too expensive). Notably, the subfield of memetic computation---where \textit{memes} are loosely defined as computationally-encoded units of task-specific information (e.g., local gradient signals) combined with global evolutionary search---offers a wealth of theories and methodologies that tackle these issues~\cite{chen2011multi,gupta2018memetic}.\\

\subsubsection{Implementation Considerations and Hardware Acceleration}

While population-based EAs offer search dynamics that complement point-based gradient descent, it is important to recognize that current algorithms may not yet provide an ``off-the-shelf" solution for PINN optimization. Indeed, a recent study based on exploratory landscape analysis shows that the features of optimization benchmarks typically used in the development of EAs fundamentally differ from those of neural network training tasks \cite{malan2025we}. Excelling at benchmarks does not, therefore, ensure effectiveness in neuroevolution. Although such in-depth analysis in the specific context of PINNs is so far unavailable, an empirical comparison of various evolution strategies (ES) by Sung \textit{et al.}~\cite{sung2023neuroevolution} shows that the performance of the classical CMA-ES \cite{hansen2001completely} is statistically similar to or better than modern variants of natural ES algorithms \cite{wierstra2014natural} as well as SGD, across a range of PINN tasks representing real-world phenomena in classical mechanics, heat and mass transfer, fluid dynamics, and wave propagation. A compelling reason to consider ES is the induced Gaussian-smoothing of loss functions, biasing flat minima that often depict better generalization performance as per the minimum description length principle \cite{hochreiter1997flat}. All algorithms tested in \cite{sung2023neuroevolution} were implemented in the JAX framework~\cite{tang2022evojax}, providing orders of magnitude faster convergence than standard implementations. The source code has been made publicly available\footnote{Python source codes for neuro-evolving PINN using EvoJAX are available at https://github.com/nicholassung97/Neuroevolution-of-PINNs.} to facilitate deeper analysis and development of customized EAs for PINNs.

Not only is there a plethora of EAs to choose from, effective application of an EA to any problem also involves careful tuning of numerous hyperparameters (e.g., learning step-size, population size, crossover rates). Interestingly, the theory of information-geometric optimization (IGO) \cite{ollivier2017information} provides a mathematical unification of probabilistic model-based EAs---including CMA-ES, natural ES, OpenAI-ES \cite{salimans2017evolution}, or techniques that replace the Gaussian model with a deep generative model \cite{hartl2024heuristically}---such that even a complete EA can be thought of as just a specific hyperparameter setting of the general IGO update rule. Empirical comparisons of optimizers suggest that more general update rules, if appropriately configured, should not underperform the ones they can approximate \cite{choi2019empirical}. It is therefore fortunate that there exists a valuable body of literature on the automatic hyperparameter configuration of EAs~\cite{tessari2022reinforcement} that could be leveraged for PINN optimization problems of varying scales and characteristics. Examples include but are not limited to adaptive population sizing \cite{nishida2018psa} and reinforcement learning for online control of learning step-sizes \cite{ nomura2025cma} or crossover rates \cite{tessari2022reinforcement}.

Another major challenge in the application of EAs is that they may scale poorly with the dimensionality of the optimized parameters. This number can indeed grow large for deep neural networks, and theoretical analysis suggests that the required number of optimization steps scales linearly with dimensions for general non-smooth functions \cite{nesterov2017random}. However, OpenAI's seminal paper on ES \cite{salimans2017evolution} prominently demonstrates that, in practice, these algorithms effectively exploit key informative dimensions of the optimization problem. This allows their performance to scale with the (low) intrinsic dimensionality of the task, rather than with the full dimensionality of the search space. This observation has been supported by recent work as well \cite{wei2025evolvableconditionaldiffusion}, positioning ES as a scalable option for training large PINNs given the availability of distributed hardware on which population evaluations can be parallelized. For tasks where the intrinsic dimensionality remains high, the latest methods of problem decomposition, cooperative coevolution, and memetic algorithms are worthy of exploration \cite{omidvar2021review}, though these have yet to be investigated in the context of PINNs.

\subsection{Neural Architecture Search for Inducing Physics Biases} \label{sec:pine-nas}

The choice of architectures and hyperparameters for PINNs has often been guided by experiential priors drawn from standard deep learning. However, a growing body of literature shows that accurate physics-informed learning may call for bespoke architectures. For example, contrary to popular belief, more hidden layers do not necessarily mean better performance and can sometimes be detrimental. An automated neural architecture search via a bilevel programming formulation revealed that, for certain governing differential equations, a shallow neural network with many neurons was more suitable for PINNs \cite{wang2024pinn}. Similar intriguing insights regarding the choice of activation function have also been uncovered. While squashing functions (e.g., \emph{tanh} and \emph{sigmoid}) and variants of the rectified linear unit are the mainstay in the deep learning literature, the use of Fourier features (i.e., \emph{sine} and \emph{cosine}) has gained prominence in physics tasks \cite{wong2022learning, jin2024fourier}, specifically for their ability to overcome spectral bias and capture periodic or high frequency patterns (e.g., shock fronts) common in various physics applications \cite{zhang2024application,sallam2023use,huhn2023physics}.

Given the large number of design choices for network architectures and hyperparameters, manual experimentation to identify an optimal design for every physical problem is impractical. Notably, there exists a rich history of research in neuroevolution that automates such design choices through the joint evolution of neural network topologies along with their parameters \cite{stanley2019designing,liu2021survey}. One such example of the fully automatic design of PINNs by a genetic algorithm with specially tailored selection, crossover, and mutation operators was proposed in \cite{kaplarevic2023identifying}. Each chromosome of this method encodes a learning algorithm, architecture of the PINN model, activation function in the hidden layers, and the output activation. In a related approach, an asynchronous genetic algorithm that alternates between model selection and optimization of physics-informed neural state space dynamics models was presented in \cite{skomski2021automating}. Daw \textit{et al.}~\cite{daw2022rethinking} proposed an evolutionary sampling procedure to incrementally accumulate sample points in regions of high PDE residuals for mitigating training failures.

In order to achieve higher approximation accuracy and faster convergence rates, an EA for concurrently exploring optimal shortcut connections between PINN layers and novel parametric activation functions expressed by binary trees was put forward in \cite{zhang2024discovering}. Interestingly, numerical studies show that random projection PINNs---where most network parameters are randomly initialized and fixed, with training limited to the output layer---significantly accelerate training on physics-informed learning objectives \cite{dong2021local}. However, ensuring predictive accuracy in this setting requires careful configuration of the randomization bandwidth: a hyperparameter defining the width of the probability distribution from which the random parameters are sampled. In \cite{dong2022computing}, this hyperparameter optimization was carried out by the differential evolution algorithm \cite{storn1997differential}, resulting in PINNs that outperformed higher-order finite element methods (an established numerical method for solving differential equations) in terms of speed and accuracy.

Training a PINN model can be computationally intensive, rendering the evaluation of hyperparameters or network architectures prohibitively expensive. To address this problem, various surrogate-assisted evolutionary techniques~\cite{jin2018data} have been proposed within the field of evolutionary computing, emerging as promising solutions. These techniques encompass a wide range of optimization scenarios, including single-objective~\cite{wang2017committee}, multi-objective~\cite{wang2021choose}, large-scale~\cite{sun2017surrogate}, constrained~\cite{wang2018global}, discrete variable ~\cite{wang2018random}, and even mixed-variable optimization~\cite{liu2021multisurrogate}. These techniques provide a strong foundation for developing more efficient hyperparameter optimization and architecture search methods tailored to PINNs.

%The use of EAs for optimizing hyperparameters balancing trade-offs between data and physics loss terms in PINN training is yet another important application. Since this is intrinsically a multi-objective optimization problem, associated discussions are contained in Section \ref{sec:pareto-optimization}.

\begin{figure*}[htbp]
\centering
\includegraphics[width=1.0\linewidth]{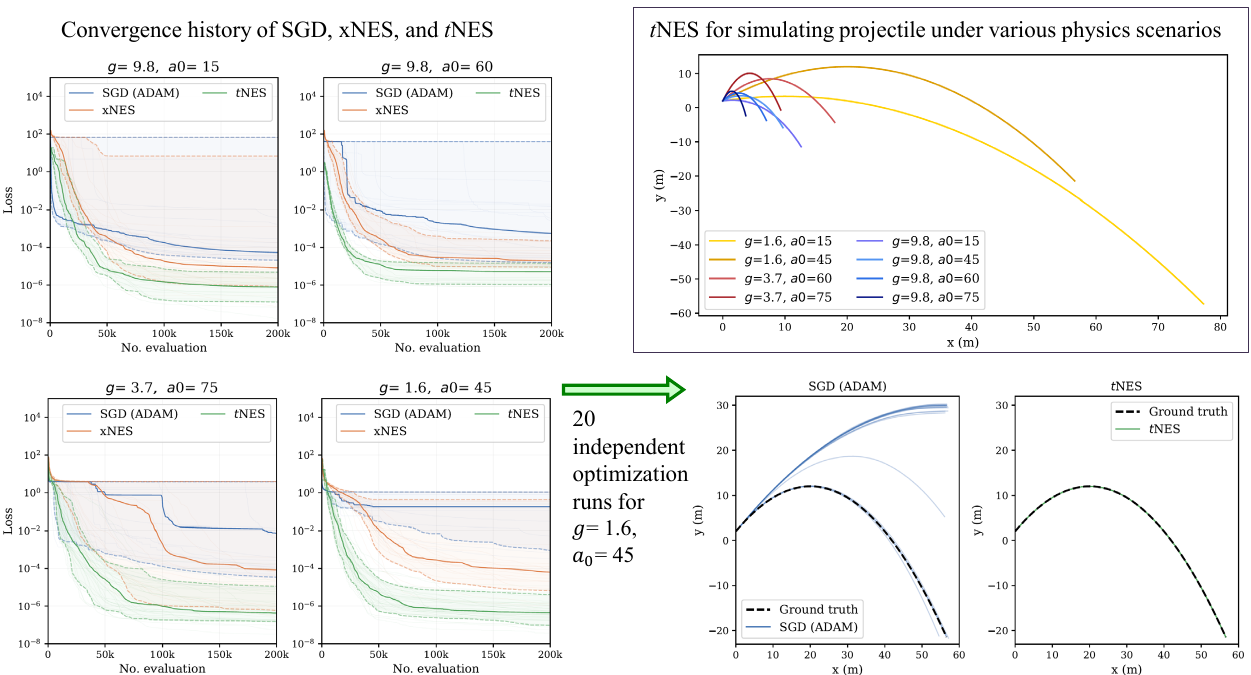}
\caption{A PINN simulating projectile motion under gravitational forces on Earth, Mars, or the moon is optimized by transfer natural evolution strategies---dubbed \textit{t}NES. The empirical comparisons of convergence trends between SGD, xNES, and \textit{t}NES  for selected scenarios are provided (the bold lines indicate the mean convergence path of multiple independent runs and the shaded areas indicate their 10th-90th interpercentile ranges). Individual solutions given by multiple SGD and \textit{t}NES (transfer from a relevant prior, i.e., source distribution obtained by solving a different gravity $g$ or launch angle $a_0$) optimization runs for one particular scenario ($g$=1.6, $a_0$=45) are plotted and compared against the ground truth. \textit{Adapted with modifications from~\cite{wong2021can}. Original figure available at [arXiv:2101.01998]}.}
\label{fig:pine-transferea}
\end{figure*}

\subsection{Transfer Optimization for Accelerated PINN Training}

\subsubsection{Learning from Related Tasks}

In a scientific study, it is common for a single set of PDEs to be evaluated under different physics scenarios (tasks), e.g., variations in PDE parameters, and IC/BCs. Moreover, for dynamical problems with causality, PINNs can generally be formulated to predict sequences of physical states, such as at different crack steps in a crack propagation study~\cite{goswami2020transfer} or at different time steps in a time-varying dynamical problem~\cite{mattey2022novel,tang2022transfer}. Such PINN models are sequentially learned using solutions from previous steps as IC/BCs. Relevant task solving experiences (i.e., optimizing PINN model parameters) can naturally accumulate over time. This allows transfer learning and optimization algorithms to exploit reusable knowledge from past episodes and jump-start the physics-informed learning, hence achieving rapid convergence for new target task~\cite{prantikos2023physics}. The transfer learning of PINN has also been shown to improve solution accuracy on challenging tasks, such as high-frequency and multi-scale problems, compared to learning the model from scratch~\cite{mustajab2024physics}. %from simpler tasks 

Transfer learning in the context of PINN commonly involves transferring network weights from a similar physics scenario that has been previously learned (source task). These transferred weights serve as initialization for training the target task. Typically, only the last few layers of the networks are optimized after the transfer, allowing the sharing of previously learned useful features between target and source tasks. A singular value decomposition-based transfer approach has also been proposed in~\cite{gao2022svd}, which involves freezing the singular vectors of the dense kernel matrix of the hidden layers and only optimizing the singular values for the target task. Apart from network weights, the study from~\cite{xu2023transfer} has shown that transfer learning can also facilitate the learning of loss-balancing weights in both forward and inverse PINN models.

Another interesting approach worth exploring is to combine physics-informed learning and data-driven learning techniques using transfer learning. A multi-fidelity transfer learning framework was proposed in~\cite{chakraborty2021transfer} to first train a PINN model to capture important information about the governing physics of the problem from the approximated (low-fidelity) physics equations. The model then undergoes data-driven learning with available high-fidelity data, leading to accurate predictions even with limited high-fidelity training data. While the study has demonstrated that transfer learning from physics can benefit a data-driven DNN, the reverse pathway deserves further investigation. There is potential for improvements to PINN optimization by harvesting the experience gained from a data-driven DNN model through transfer learning and transfer evolutionary optimization. \\

\subsubsection{Adaptive Transfer}

An important question in the transfer of physics-informed learning is the identification of relevant source task(s). If the source task is far from the target task, there could be insufficient commonalities between these tasks for the transfer to be beneficial, and it could even negatively impact the learning performance of the target task (negative transfer). For problems with causality, a natural choice for \textit{source} is the networks learned on the previous time-step or physical state~\cite{mattey2022novel,tang2022transfer}. However, the target function is generally \textit{a priori} unknown in physics-informed learning, hence calculating the similarity between target and source tasks based on their output functions is not feasible. Identifying the most relevant source task then becomes a challenge that requires prior knowledge of how the system behaves when its geometry, PDE, and IC/BC parameters change~\cite{liu2023adaptive}. This becomes even more challenging when dealing with previously solved tasks involving multiple parameter changes.

Adaptive transfer learning can be a particular strategy to address the issues without the need for explicit \textit{a priori} identification of relevant sources. For example, an adaptive transfer physics-informed learning framework was proposed in~\cite{liu2023adaptive} to achieve effective transfer, even when there is significant variation in the PDE parameters between the source and target tasks. The approach treats the PDE parameters as trainable, hence, they can be adaptively updated from the source to target parameter values, creating a low-loss path in the parameter space to guide the optimization from source to target tasks. Throughout this process, the loss is consistently kept at a low level (similar to the concept of a minimum energy path connecting two minima~\cite{garipov2018loss}), ensuring that the PINN solution remains very close to the physical solution and guaranteeing the stability of the training process.

A recent study by Wong \textit{et al.}~\cite{wong2021can} on the transfer neuroevolution of PINN demonstrated that accurate solutions for new and harder PDEs can be rapidly achieved by optimizing PINN with EA. A key aspect of this transfer optimization approach is its ability to incorporate multiple sources of knowledge while optimizing the PINN for the specific target task. This can be achieved through a principled probability mixture model-based EA that adaptively assigns weights to the different sources based on their observed relevance to the target task during training, thereby preventing detrimental negative transfers. Fig.~\ref{fig:pine-transferea} illustrates the accelerated convergence trends achievable through the transfer neuroevolution of PINN, with clear optimization advantages relative to xNES (a state-of-the-art of EA~\cite{wierstra2014natural}) and SGD-based (Adam) training.

Similar probabilistic formulations are also possible in multitask neuroevolution~\cite{gupta2022half, zhang2022multitask}, enabling the simultaneous solving of sets of related PDEs in a single algorithmic pass. A multitask approach could prove much faster than solving PDEs independently, as useful information is shared between tasks without the need to optimize from scratch.

\begin{figure*}[htbp]
\centering
\includegraphics[width=1.0\linewidth]{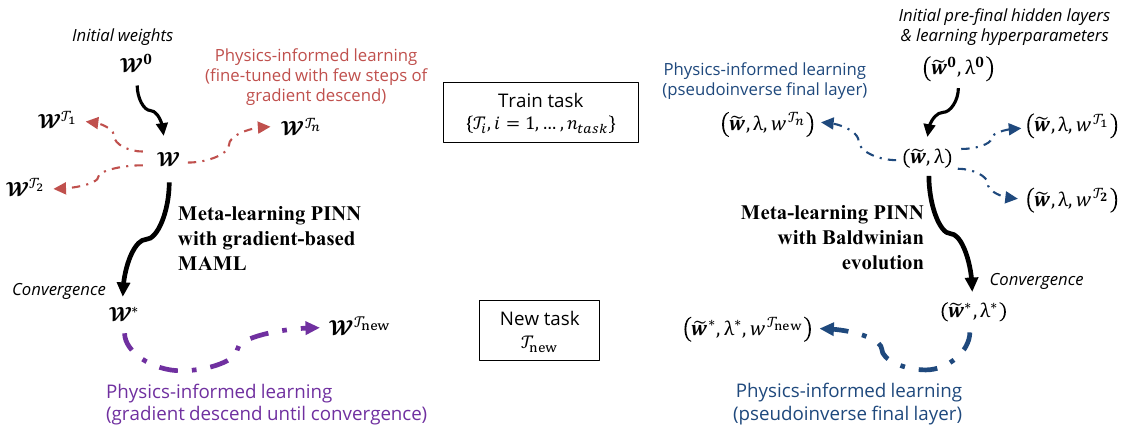}
\caption{Meta-learning PINN with MAML (left) versus Baldwinian evolution (right). In MAML, the initial PINN weights are learned using gradient-based method, such that they can be quickly fine-tuned (physics-informed learning) on new tasks. The task-specific fine-tuning is limited to one or a few gradient descents during training, which is usually insufficient for a meta-learned PINN to converge. In Baldwinian evolution, the weight distribution in the pre-final nonlinear hidden layers $\vb*{\tilde{w}}$ and learning hyperparameters $\lambda$ are jointly evolved. The task-specific fine-tuning is performed on the output layer weights $w$ (segregated from $\vb*{\tilde{w}}$) with a 1-step pseudoinverse operation at both training and test time (for linear ODE / PDEs). \textit{Adapted with modifications from~\cite{wong2024generalizable}. Original figure available at [arXiv:2312.03243]}.}
\label{fig:pine-meta-learning}
\end{figure*}

\subsection{Evolutionary Meta-Learning for Generalizable PINNs}

\subsubsection{Meta-Learning in PINNs}

Meta-learning algorithms provide a systematic framework for reusing knowledge from a distribution of learning tasks, allowing models to quickly adapt to new physics-informed learning tasks \textit{with minimal training}. Crucially, meta-learning offers a pathway towards a generalizable PINN over a broad class of physics scenarios, including different PDE families and geometries, as highlighted in Subsection~\ref{sec:pinn-generalize}. For example, the MAML and Reptile algorithms have been recently explored in the context of PINN~\cite{penwarden2023metalearning,cho2024hypernetwork,liu2022novel}. They seek to meta-learn an optimal model initialization in the space of network weights from which the PINN can rapidly converge to a solution that minimizes the physics-informed loss of new tasks. This is achieved by a task-specific fine-tuning procedure during the meta-learning stage through an inner-outer loop training framework. This problem can be mathematically posed as a two-stage or bi-level optimization. The network weights are trained explicitly (in the outer loop) by optimizing the model performance over a batch of physics-informed learning tasks after one or a few gradient descent updates (at the inner loop) from the learned initialization.\\

\subsubsection{Neural Baldwinism}

\begin{figure*}[htbp]
\centering
\includegraphics[width=1.0\linewidth]{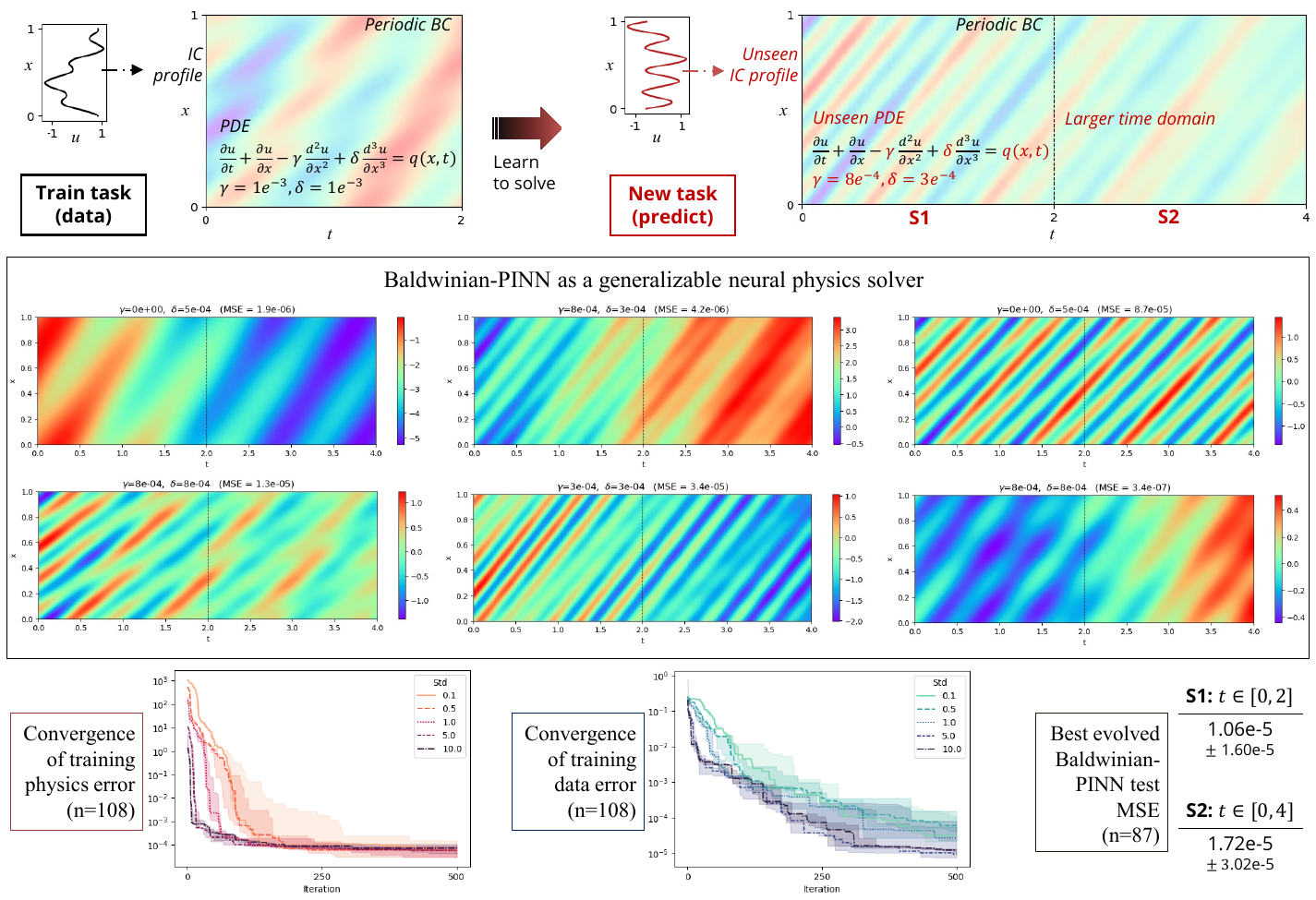}
\caption{Schematic to illustrate new tasks arising from \textit{family of PDEs problem}: \textbf{S1} change to new PDE ($\gamma$ and $\delta$ parameters), source term $q$, and IC profile for $t\in[0,2]$ (same time domain as train tasks), and \textbf{S2} projection to longer time domain $t\in[0,4]$. The solution for new PDE tasks (they are visually indistinguishable from the ground truth) can be quickly obtained in the order of milli-seconds by the Baldwinian-PINN~\cite{wong2024generalizable}. In~\cite{wong2024generalizable}, CMA-ES was employed to evolve the Baldwinian-PINN models, demonstrating effective convergence in both physics residuals and MSE (108 training tasks). Superior performance is achieved with CMA-ES using initial standard deviations of 5 and 10. The mean MSE across 87 test tasks is below 5e-5.\textit{Adapted with modifications from~\cite{wong2024generalizable}. Original figure available at [arXiv:2312.03243]}.}
\label{fig:pine-metaea}
\end{figure*}

\begin{figure*}[htbp]
\centering
\includegraphics[width=1.0\linewidth]{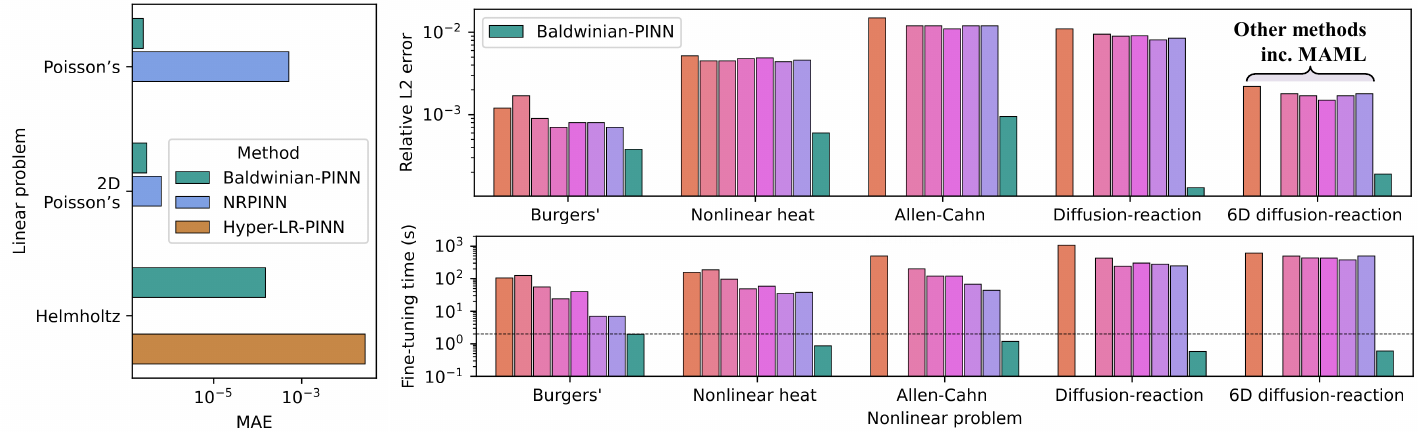}
\caption{Empirical comparison between results reported in Baldwinian-PINN~\cite{wong2024generalizable} and recent meta-learning PINN studies. For linear problems, the results for NRPINN and Hyper-LR-PINN are extracted from~\cite{liu2022novel} and~\cite{cho2024hypernetwork}, respectively. For nonlinear problems, the results for other meta-learning PINN models / methods (inc. MAML) are extracted from~\cite{penwarden2023metalearning}. On nonlinear test tasks, evolved Baldwinian-PINNs achieve up to 10x lower relative norm error, with prediction times under 2 seconds---over 100x faster than other meta-learning methods. For instance, in diffusion-reaction experiments, Baldwinian-PINN yields 70x better accuracy with 700x less computation time.}
\label{fig:pine-metaea-result}
\end{figure*}

Recently, a meta-learning PINN framework based on the evolutionary principle of Baldwinism---dubbed \textit{Baldwinian} PINN~\cite{wong2024generalizable}---has been proposed to address the training challenges faced in conventional gradient-based meta-learning algorithms such as MAML. The \emph{Baldwin effect} in nature explains the development of phenotypic plasticity among a population's individuals \cite{downing2012heterochronous}, whereby their young are “born ready” with strong learning biases to perform a wide range of tasks. In order to pre-wire such learning ability into the initial connection strengths of a neural network, enabling it to be performant over a distribution of physics-informed learning tasks, an algorithmic realization of neural Baldwinism is examined in the context of PINNs. The algorithm tackles the two-stage optimization problem in meta-learning by coupling hidden layer evolution, for better generalization across tasks, with lifetime-learning at the output layer to specialize to particular tasks. Task-specific learning is limited to the output layer only~\cite{zou2023hydra,desai2021one}, allowing the model to solve new linear PDE problems with a one-step pseudoinverse operation (or a few pseudoinverse steps for nonlinear PDE problems). The pseudoinverse operation has a closed form and is therefore fast. Unlike gradient-based MAML, the global evolutionary search in Baldwinian-PINNs is better equipped to handle the two-stage/bi-level problem formulation where exact gradients may be difficult to compute. Being a gradient-free algorithm, it does not require unrolling and differentiating through the computation graph, making Baldwinian-PINN an efficient method that scales well with the number of GPUs available. Fig.~\ref{fig:pine-meta-learning} illustrates the contrast between MAML and Baldwinian evolution for PINN\footnote{Python source codes for Baldwinian-PINN are available at https://github.com/chiuph/Baldwinian-PINN.}. 

Experiments in \cite{wong2024generalizable} clearly illustrate how Baldwinian-PINN can consistently solve new physics scenarios within a diverse task distribution with orders of magnitude improvement in speed and accuracy compared to a single PINN model. These PDEs include both linear equations such as the convection-diffusion, Helmholtz and Poisson equations, and non-linear equations such as the Burger's, Allen-Cahn and diffusion-reaction equations. Examples of effective generalization of a Baldwinian-PINN to an entire PDE family with diverse output patterns is demonstrated in Fig.~\ref{fig:pine-metaea}, on two task scenarios: \textbf{S1} shows successful learning of solution for unseen set of PDEs, including changes to PDE parameters, source term, and IC profile; while \textbf{S2} shows effective extrapolation of solution to a longer time domain. Baldwinian-PINNs have clear benefits for the continual modeling of dynamical systems through their versatility in handling ICs and ability to rapidly model and stitch time windows together with minimal error. Crucially, Baldwinian evolution does not require explicit parameterization, and evolved Baldwinian-PINN can flexibly generalize to new ICs, BCs, and PDE source terms, a stark contrast to the need for existing parameterized PINNs to have pre-defined PDE, BC or IC parameters. Overall, the outstanding results in \cite{wong2024generalizable} compared to other meta-learning methods (Fig.~\ref{fig:pine-metaea-result}) strongly motivate similar approaches in the spirit of Baldwinian-PINN.\\
%Interestingly, the Baldwinian-PINN maintain good accuracy on tasks from \textbf{S2}, whereby the model first learns the PDE solution for $t\in[0,2]$, before using the learned solution $u(x,t=2)$ as new IC for solving $t\in[2,4]$. as they require interpolation across a potentially infinitely large distribution of tasks (e.g. possible BCs or ICs), further emphasizing the merits of Baldwinian evolution for physics. 

\subsubsection{Meta-Learning Neural Architecture, Loss Function, and Optimizer}

Most gradient-based meta-learning PINN approaches proposed in the literature aim to find an optimal initialization of the network weights. In the study by~\cite{psaros2022meta}, a meta-learning framework for the PINN loss function that incorporates the underlying physics of a parametrized PDE has demonstrated significant performance improvement for new tasks. In~\cite{bihlo2024improving}, the complexity of the PINN optimization problem was proposed to be tackled by \emph{learning} a customized optimizer that's itself parameterized by a neural network. Notably, the learning of the optimizer was done using a persistent ES algorithm \cite{vicol2021unbiased}, and showed that meta-learned optimizers could substantially improve vanilla PINNs over standard hand-crafted optimizers such as Adam. %by reaching particular error levels quicker than

There are already a handful of studies that have proposed using EAs to find performant neural architectures and hyperparameters for a single PINN (ref. to Subsection~\ref{sec:pine-nas}). There is huge scope for future research and development into using EAs to jointly meta-learn model initialization, neural architecture, loss function (hyperparameters), and even the gradient-based optimizer to drastically improve convergence on a set of PINN tasks. For example, an EA can jointly optimize the neural architecture and network weights for a PINN such that the meta-learned PINN model utilizes physics compliance during prediction of a new scenario to enhance generalization across diverse tasks with varying initial conditions, PDE parameters, and geometries.

There is significant potential in exploring evolutionary computation methods to develop meta-learnable optimizers, loss functions, neural architectures, and initializations tailored to diverse physics-informed learning tasks, thereby presenting opportunities to enhance the generalizability of PINN models. Lastly, we would like to make additional remarks on the important role of labeled data in meta-learning PINN. In general, the meta-learning incurs a higher upfront training cost to achieve good generalization. Some proposed methods such as the Baldwinian-PINN cleverly utilize a data-centric meta-learning objective, i.e., using data MSE rather than purely physics-driven loss. Meta-learning PINN algorithms can be meaningfully guided by a few tasks with labeled data, easing optimization difficulties caused by the complex loss landscape defined by the physics-informed loss. \\

\subsubsection{Neural Operator Networks}

A notable alternative to learning general PDE solutions is the use of neural operator networks \textit{in a purely data-driven setting}~\cite{goswami2023physics, azizzadenesheli2024neural, cao2024fast}. By learning mappings between infinite-dimensional function spaces, neural operators enable prediction of solutions in real-time under different PDE source terms, ICs or BCs. For instance, consider the 1D Poisson equation $\frac{d^2u}{dx^2} = f(x)$; a neural operator aims to learn the mapping $G: f \to u$ from the source term $f(x)$ to the PDE solution $u(x)$. Prominent neural operator architectures encompass the Deep Operator Network (DeepONet)~\cite{lu2021learning} and the Fourier Neural Operator (FNO)~\cite{li2024physics}. DeepONet comprises a branch network to encode the input function at predetermined sample locations, and a trunk network to encode spatial coordinates at which the output function is evaluated. Conversely, FNO employs spectral convolutions to efficiently model nonlocal interactions and directly predict the entire field solution, making it particularly suited for structured-grid data~\cite{li2023fourier}. Further extensions, such as graph-based neural operators~\cite{kovachki2023neural}, transformer operators~\cite{hao2023gnot}, and geometry-informed operators~\cite{li2023geometry}, broaden this framework to manage irregular domains and bolster computational efficiency. Similar to other data-driven methods, neural operators are inherently data-intensive, typically necessitating large PDE simulation datasets to achieve robust generalization.

The integration of physics-informed losses, such as PDE constraints, into neural operators can substantially reduce training data requirements by leveraging known physics~\cite{wang2021learning}. This approach establishes PINNs and neural operators as complementary techniques. Nonetheless, analogous to parameterized PINNs, adherence to physical laws during training does not necessarily translate to their preservation in out-of-sample predictions, thereby presenting common generalization challenges. As a result, neural operators often resort to using larger training datasets to enhance generalization. Furthermore, unlike parameterization-agnostic Baldwinian-PINNs, they demand explicit parameterization of the conditions under which the model is intended to generalize. Although this explicit requirement diminishes the flexibility of neural operators relative to meta-learning strategies, they remain a vital complementary method for generalized neural PDE solvers. Currently, the effectiveness of EAs for optimizing neural operators remains an open area of research. Early work bringing randomization into shallow operator networks has appeared in~\cite{fabiani2025randonets}, showing their potential to outperform standard operator networks in both approximation accuracy and computational efficiency. %Additionally, DeepONet has been successfully integrated with genetic algorithms for inverse design applications in material science, underscoring the promising synergy between neural operators and evolutionary computation~\cite{lu2022multifidelity}.

%\textcolor{blue}{A notable alternative to generalized PINNs is through neural operator networks \cite{goswami2023physics, li2024physics, cao2024fast}, which, by learning mappings between infinite-dimensional function spaces, allow models to solve PDEs in real-time under many different ICs or BCs, requiring only light task-specific fine-tuning. However, unlike parameterization-agnostic Baldwinian-PINNs, neural operators call for explicit parameterization of the conditions over which the model is to generalize. Although this does make neural operators less flexible compared to meta-learning, they are considered an important complementary approach to model generalization. At the time of writing of this paper, the effectiveness of EAs for optimizing physics-informed neural operator nets remains an open question. Some foundational work bringing randomization into shallow operator networks has appeared in \cite{fabiani2025randonets}, showing their potential to outperform standard operator nets by several orders of magnitude in terms of approximation accuracy and computational cost.}

\subsection{Evolving Interpretable Models Beyond PINNs}

In scientific machine learning, a key objective of discovering new insights from experimental data relies on human-interpretable models or outputs. While advances have been made in explaining the outputs of deep neural networks, these models are largely perceived as black-boxes \cite{ali2023explainable}. The same issue naturally afflicts PINNs, prompting the brief discussion below on the evolution of other model types.

In contrast to neural networks, the subfield of \emph{genetic programming} in evolutionary computation has a long history of successfully discovering free-form natural laws from data through symbolic regression \cite{schmidt2009distilling, dong2025recent}. Recently, these algorithms have even been used for solving high-dimensional PDEs, with the distinct advantage of offering compact analytical expressions approximating complex solutions~\cite{cao2024interpretable, cao2023genetic, cao2025interpretable}. Balancing compactness and model accuracy is a non-trivial task that is well-suited to evolutionary multi-objective optimization. In addition to symbolic regression, a highly interpretable mathematical description of dynamical systems can be obtained by identifying a best-fit linear dynamics model (i.e., a linear system of governing ODEs) from data. Although finding such a data-driven model that simultaneously satisfies physics constraints translates into a highly constrained, nonlinear and non-convex optimization problem \cite{askham2018variable}, it has been shown to be efficiently solvable by evolution strategies \cite{gupta2024globally}. Adaptations of this evolutionary approach to identify nonlinear dynamics have also been explored \cite{hazelden2023evolutionary}, with the final prediction accuracy and training stability of evolution exceeding those of models trained by gradient descent. %, through neural ODEs \cite{chen2018neural},

\section{Conclusion}

As the influence of deep learning permeates beyond typical natural language processing or computer vision tasks, it is clear that rationalizable models that adhere to the fundamental laws of nature are a need of the hour. Instilling centuries of accumulated wisdom, in the form of mathematically expressible scientific knowledge, into otherwise naive statistical models is crucial. Research efforts in the past few years have demonstrated that PINNs are a promising approach to that end. 

However, like any new technique, PINNs inevitably call for new technical innovations. Many of the approaches commonly used in training standard data-driven neural networks need to be re-evaluated, such as the method of network initialization and the choice of activation functions. The use of purely gradient-based optimization, which includes the hugely successful family of SGD algorithms, is especially worthy of reconsideration in the light of analyses summarized in this work. Identified challenges pertaining to local minima traps, fixed point traps, and vanishing gradients of physics-informed learning objectives highlight how these can prove deleterious for PINNs. These obstacles to conventional gradient-based optimization serve to outline ways in which gradient-free algorithms of computational intelligence may be uniquely advantageous as substitutes for or in synergy with SGD. Indeed, the crafting of synergistic optimization frameworks that meld gradient-based and gradient-free optimization unveils novel research questions and directions, firmly positioning scientific machine learning as an emerging area of translational interest for the computational intelligence community.

\section*{Acknowledgment}
This research was in part supported by the National Research Foundation, Singapore through the AI Singapore Programme, under the project ``AI-based urban cooling technology development" (Award No. AISG3-TC-2024-014-SGKR), in part by the National Research Foundation, Singapore and DSO National Laboratories under the AI Singapore Programme - ``Design Beyond What You Know: Material Informed Differential Generative AI (MIDGAI) for LightWeight High-Entropy Alloys and Multi-functional Composites (Stage 1b)" (Award No. AISG2-GC-2023-010). Abhishek Gupta is supported by the Ramanujan Fellowship from the Anusandhan National Research Foundation, Government of India (Grant No. RJF/2022/000115).

\ifCLASSOPTIONcaptionsoff
  \newpage
\fi



%\begin{thebibliography}{1}


\bibliographystyle{IEEEtran}
\bibliography{IEEEabrv,Bibliography}
%\end{thebibliography}


\end{document}